\documentclass[runningheads]{llncs}

\usepackage[T1]{fontenc}

\usepackage{graphicx,verbatim}

\usepackage{amsmath, amssymb} 
\usepackage{graphicx}
\usepackage{hyperref}
\usepackage{orcidlink}

\begin{document}

\titlerunning{Benchmarking Uncertainty Quantification and its Disentanglement} 
\title{Benchmarking Uncertainty and its Disentanglement in multi-label Chest X-Ray Classification}

\author{
Simon Baur\inst{1}\orcidlink{0009-0009-4307-3078}, 
Wojciech Samek \inst{1,2,3}\orcidlink{0000-0002-6283-3265},
Jackie Ma\inst{1}\orcidlink{0000-0002-2268-1690}
}

\authorrunning{Simon Baur et al.}

\institute{Fraunhofer Heinrich-Hertz-Institut, 10587 Berlin, Germany\\
\email{\{simon.baur, wojciech.samek, jackie.ma\}@hhi.fraunhofer.de}\\
\and
Technische Universität Berlin, 10623 Berlin, Germany\\
\and 
The Berlin Institute for the Foundations of Learning and Data (BIFOLD), 10587 Berlin, Germany\\
}

\maketitle              

\begin{abstract}
Reliable uncertainty quantification is crucial for trustworthy decision-making and the deployment of AI models in medical imaging. While prior work has explored the ability of neural networks to quantify predictive, epistemic, and aleatoric uncertainties using an information-theoretical approach in synthetic or well defined data settings like natural image classification, its applicability to real life medical diagnosis tasks remains underexplored. In this study, we provide an extensive uncertainty quantification benchmark for multi-label chest X-ray classification using the MIMIC-CXR-JPG dataset. We evaluate 13 uncertainty quantification methods for convolutional (ResNet) and transformer-based (Vision Transformer) architectures across a wide range of tasks. Additionally, we extend Evidential Deep Learning, HetClass NNs, and Deep Deterministic Uncertainty to the multi-label setting. Our analysis provides insights into uncertainty estimation effectiveness and the ability to disentangle epistemic and aleatoric uncertainties, revealing method- and architecture-specific strengths and limitations.

% Authors must provide keywords and are not allowed to remove this Keyword section.

\end{abstract}
\section{Introduction}
Deep neural networks have widely been used to solve medical image analysis tasks, such as image classification and segmentation \cite{Litjens2017,zhou2021review}. However, in high-stakes applications such as radiology, models must meet rigorous standards to ensure reliability and trustworthiness. These requirements are developed from multiple perspectives, including the needs of patients, medical professionals, and regulatory agencies. From a technical standpoint, comprehensive benchmarks are essential to guide the development of safe and interpretable AI models. In radiology, AI-assisted diagnosis can reach human-expert level performance \cite{Rajpurkar2021}, but the expressivity of uncertainty of such models is often not intensively studied. A false negative AI classification of pneumonia, for example, could delay life-saving treatment, while overconfident AI predictions could lead to unnecessary interventions. Regulatory agencies such as the FDA \cite{FDA_AI_Uncertainty} emphasize the importance of uncertainty-aware AI systems, yet robust uncertainty quantification remains underexplored in multi-label medical imaging \cite{leibig2017leveraging}. Despite the increasing focus in benchmarking AI models on performance, robustness, and explainability \cite{Springenberg2023,Clark2025}, uncertainty quantification (UQ)\cite{gawlikowski2023survey,abdar2021review} remains an underexplored aspect of model evaluation. Reliable UQ methods are crucial for assessing predictive confidence, and ultimately improving model trustworthiness in clinical practice \cite{abdullah2022review,evans2024understanding}. Our work addresses this research gap by systematically benchmarking a diverse set of uncertainty quantification methods for multi-label chest X-ray classification. We provide a comprehensive evaluation across convolutional and transformer-based architectures, offering practical insights into their reliability and applicability in real-world medical imaging workflows.

\subsection{Related Work}
\subsubsection{Benchmarking Uncertainty Quantification.}
\label{sec:related_work}
Our work follows the benchmarking setup recently proposed by Mucsányi et al. \cite{mucsanyi2024benchmarking} for uncertainty quantification in natural imaging, which we adapt to a multilabel setting. Their framework benchmarks UQ methods across five tasks, assessing their ability to disentangle uncertainty and evaluating various aggregation strategies.

\subsubsection{Distributional Methods.} Distributional methods model uncertainty by learning a second-order predictive distribution $p(y \mid x, \theta)$, where the model outputs a distribution over predictions rather than a point estimate. Practically, $M$ forward passes are sampled for each input $x$ to approximate $p(y \mid x, \theta)$ \cite{wang2020survey,wilson2020bayesian}. We evaluate a diverse set of uncertainty quantification methods, including Latent Heteroscedastic Classifier (HET-XL) \cite{collier2023massively}, Swag \cite{maddox2019simple}, MC Dropout (MC-D) \cite{srivastava2014dropout}, Heteroscedastic Classification Neural Network (Het-NN) \cite{collier2021correlated}, Deep Ensemble (D-Ens) \cite{lakshminarayanan2017simple}, Shallow Ensemble (SE) \cite{DBLP:journals/corr/LeePCCB15}, Evidential Deep Learning (EDL) \cite{sensoy2018evidential}, Masked Attention (M-Attn) \cite{baur2025advancing}, and Gumbel Softmax (GS-Attn) \cite{Pei2022}. The latter two are only applicable to transformers.

\subsubsection{Deterministic Methods.} Deterministic uncertainty methods estimate uncertainty without requiring a distribution over predictions. These approaches often rely on model-internal features to assess confidence in predictions. We evaluate several deterministic methods, namely Loss Prediction (LP) \cite{yoo2019learning}, Correctness Prediction (CP), Temperature Scaling (Temp) \cite{guo2017calibration} and Deep Deterministic Uncertainty (DDU) \cite{mukhoti2023deep}.

\subsubsection{Information-Theoretical Approach and Disentanglement.} The information-theoretical (IT) formulation of uncertainty \cite{depeweg2018decomposition,hullermeier2021aleatoric} models predictive uncertainty as a combination of epistemic uncertainty (EU), arising from limited data or model capacity, and aleatoric uncertainty (AU), reflecting inherent noise in the data. While this decomposition has been thoroughly investigated in controlled and synthetic settings, recent studies \cite{kahl2024values,mucsanyi2024benchmarking} raise concerns about the practical effectiveness of the IT framework in complex, real-world datasets, questioning its ability to reliably disentangle epistemic and aleatoric uncertainties, and therefore its effectiveness of modeling uncertainty in total. 

\subsection{Contribution}
Our key contributions can be summarized as follows:
\begin{enumerate}
    \item We present the first extensive empirical benchmark of 13 uncertainty methods on chest X-ray data, spanning multiple tasks and both ConvNets and Vision Transformers.
    \item We adapt three UQ methods—EDL, Het NN, and DDU—from multiclass to multilabel classification.
    \item Our analyses highlight limitations in disentangling EU and AU using the IT framework. We infer that the IT framework is to be used with caution. 
\end{enumerate}
\section{Theoretical Framework and Methods}
In radiology, epistemic uncertainty stems from limited knowledge or distribution shifts, affecting underrepresented or unseen cases and signaling the need for expert review or more data. Aleatoric uncertainty, originating from inherent data variability (e.g. motion artifacts), can be used to flag images for re-acquisition. Methods that are reliably able to distinguish EU and AU could therefore significantly advance AI deployment in diagnostic applications.

\subsection{Uncertainty Disentanglement}
\label{sec:uc_decomposition}
To derive uncertainty estimates $u(x)$ for an input $x$ we focus on the information-theoretical (IT) \cite{depeweg2018decomposition,hullermeier2021aleatoric} approach of decomposing total predictive uncertainty (PU) into aleatoric uncertainty (AU) and epistemic uncertainty (EU). The model's predictions are obtained via the Bayesian-Model-Average (BMA):
\begin{equation}
\text{BMA} \equiv \bar{\pi}(x) \equiv \frac{1}{M} \sum_{m=1}^{M} \pi^{(m)},
\end{equation}

\begin{equation}
\text{PU} \equiv \mathbb{H} \left( \bar{\pi}(x) \right),
\end{equation}

\begin{equation}
\text{AU} \equiv \frac{1}{M} \sum_{m=1}^{M} \mathbb{H} \left( \pi^{(m)} \right),
\end{equation}

\begin{equation}
\text{EU} \equiv \mathbb{H} \left( \bar{\pi}(x) \right) - \frac{1}{M} \sum_{m=1}^{M} \mathbb{H} \left( \pi^{(m)} \right),
\end{equation}

where $M$ is the number of sampled forward passes, $\pi$ is the class probability output vector and $\mathbb{H}(x)$ is the entropy.

\subsection{Multilabel Adaptions}
We adapt EDL, Het-NN and DDU, methods originally designed for multiclass tasks, to the multilabel setting. For a detailed mathematical description see \nameref{sec:appendix_a}. 
\section{Tasks, Experimental Setting and Results}
\subsection{Tasks}
We evaluate all methods mentioned in \ref{sec:related_work} on 6 different tasks on the MIMIC-CXR-JPG \cite{johnson2019mimiccxrjpglargepubliclyavailable} dataset (official train/val/test split) which is annotated for 14 pathologies with labels being either 0 (negative), 1 (positive), or -1 (uncertain). We further provide correlation analysis of aleatoric and epistemic uncertainty as well as insights into capabilities of uncertainty disentanglement of each method. 

\subsubsection{Task 1: OOD-Detection.} 
Out-of-distribution detection assesses a model's ability to capture epistemic uncertainty. We construct an OOD dataset by randomly sampling unaltered images from MIMIC-CXR-JPG (assigning label 0, marking ID data) and applying a combination of transformations (Gaussian noise, motion blur, vignettes, mask occlusion, etc.) to an equal number of images (assigning label 1, marking OOD data). These transformations induce a performance drop of $\sim$ 20\% on validation AUROC, similar to the difference between ImageNet and ImageNet-C. OOD detection performance is evaluated using binary AUROC, where a higher score reflects better separation between ID and OOD samples.

\subsubsection{Task 2: Uncertainty Label Prediction.} 
 MIMIC-CXR-JPG includes class labels for each class annotated as negative (0), positive (1), or uncertain (-1), where the latter is therefore providing a source of expert-labeled ground truth uncertainties. To assess aleatoric uncertainty, we evaluate how well the model's uncertainty scores align with these ground truth uncertainties. Specifically, we filter the test set to include only samples with at least one uncertain (-1) label across all classes, mapping -1 to 1 (effectively marking the class as uncertain) and both 0 and 1 to 0 (effectively marking the class as certain). We then compute the binary AUROC per class and report the macro-average AUROC, where higher values indicate better identification of ground truth uncertainties.

\subsubsection{Task 3: Correctness Prediction.} 
To assess the model's ability to judge its own prediction reliability, we infer correctness labels through assigning 1 to correct predictions (threshold 0.5 for positives) and 0 to incorrect ones. We then compute binary AUROC per class between estimated UQ scores and correctness labels, and report the macro-average AUROC, where a higher score indicates better distinction between correct and incorrect predictions.

\subsubsection{Task 4: Abstained Prediction.}
To assess the impact of uncertainty on accuracy, we iteratively remove the 5\% most uncertain samples (as determined by estimated UQ scores) and track accuracy, effectively making the test set more certain in each iteration. We compute the Area Under the Accuracy Coverage Curve (AUAC), where higher values indicate better uncertainty-informed abstention.

\subsubsection{Task 5 and Task 6: Calibration.} 
For calibration assessment, we compute Expected Calibration Error (ECE) and Maximum Calibration Error (MCE) per class. We report the macro-average ECE and MCE across all classes, where lower values indicate better calibration.

\subsection{Experimental Setting.}
We trained each model for 50 epochs with early stopping of 5 based on validation AUROC. Learning rate was initialized with 1e-4 and decreased with a cosine annealing learning rate scheduler. We used the AdamW optimizer with a weight decay of 0.1. We trained five models per method with different seeds. For distributional methods, we performed M=5 forward passes per model, totaling 25 samples per input. We report the mean and standard error across all models and forward passes. We chose $M=5$ to keep our experiments computationally efficient and because higher $M$s did not show significant improvements \cite{mucsanyi2024benchmarking}. We chose ViT-tiny and ResNet18 models due to their comparable parameter sizes, not utilizing larger models as they yielded only minor performance improvements on validation metrics. We initialized all models with weights pretrained on ImageNet. We retained method-specific hyperparameters as reported from the best-performing runs in \cite{mucsanyi2024benchmarking}. 

\subsection{Results}

\begin{figure}[!h]
\includegraphics[width=\textwidth]{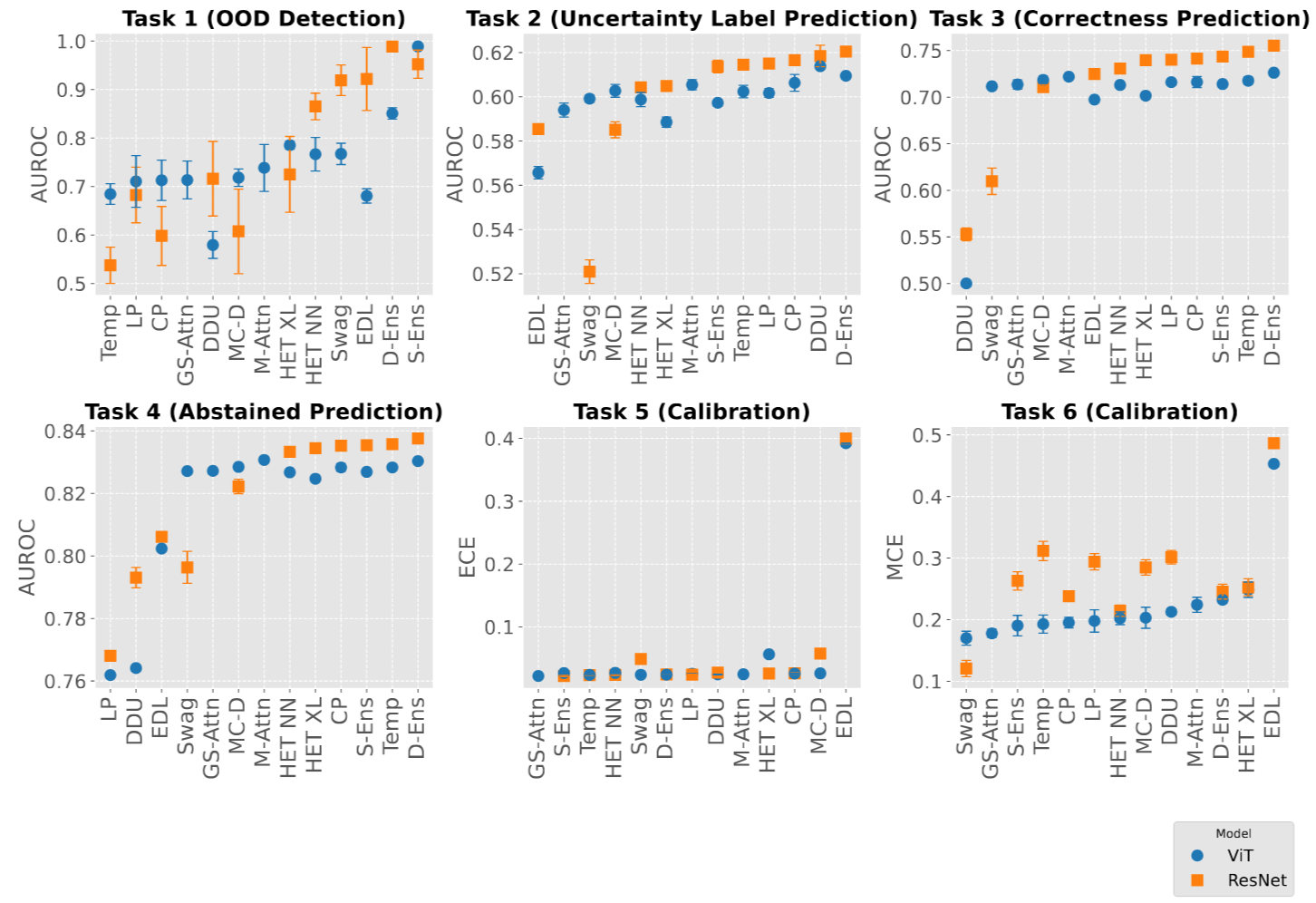}
\caption{Results for Tasks 1-6: ViT-Tiny (blue) and ResNet-18 (orange). For Tasks 1–4, we computed three uncertainty scores per method—predictive (PU), epistemic (EU), and aleatoric (AU)—as defined in~\ref{sec:uc_decomposition}. Each score yields a separate AUROC, and we report only the highest per method. Task 5 and 6 are evaluated based on BMA scores.} \label{fig_3:method_results}
\end{figure}

Fig.\ref{fig_3:method_results} summarizes results across Tasks 1-6 and reports the best performing of PU, AU and EU scores for each method. Task 1 exhibits a wide performance range (0.55–0.99), implicating a strong impact of method choice on OOD detection. For ViT, S-Ens and D-Ens perform best (only two methods achieving > 0.9), while ResNet benefits most from S-Ens, D-Ens, EDL, and SWAG, all achieving scores above 0.9. Distributional methods in general are outperforming deterministic methods across model architectures. Task 2 scores are varying less (almost all methods scoring between 0.58–0.62) compared to Task 1, with D-Ens and DDU being the top performing. Swag on ResNet is an outlier towards the bottom (0.52), indicating poor capability of capturing aleatoric uncertainties. Notably more deterministic methods lie within the upper performance range compared to Task 1. Task 3 again displays a narrow range for most methods (0.7-0.75). Again, Ensemble based methods (D-Ens, S-Ens) are among top scorers. CP, which specifically is designed to solve Task 3 shows high performance, and therefore aligns well with it's intention. DDU and Swag for ResNet show significantly lower performance than other methods here. Task 4 paints a similar picture as Task 3 (most ranging from 0.82–0.84), this time lower outliers being LP (both) and DDU (ViT), indicating weak alignment with accuracy for those methods. Again, D-Ens and S-Ens are top scorers. Results for Task 5 suggest almost all methods are well calibrated w.r.t ECE,(ECE < 0.04), except EDL (ECE = 0.4) which is strongly ill-calibrated for both models. Calibration in terms of MCE paints a more diverse picture: for ViT, most methods align around 0.2, whereas ResNet shows strong differences, with Swag being best (0.12) and almost all other methods displaying a MCE > 0.25. EDL is again heavily miscalibrated. Results across both calibration tasks therefore imply that swag is leading across calibration metrics and architectures, while EDL has significant trouble in achieving good calibration. 

We now go on to analyze the promised disentanglement capabilities of the IT approach. As a first step we investigate rank correlations of EU and AU scores on Task 1 and 2, which represent tasks aimed to be solved by the respective scores. If the promise of the IT approach holds, EU and AU scores should not display significant correlations. However, almost all methods show significant correlations between EU and AU. See \nameref{sec:appendix_b}  for detailed correlation metrics. We therefore can confirm prior findings of e.g. \cite{mucsanyi2024benchmarking} that UQ scores as disentangled by the IT approach do not keep their promise of uncorrelated uncertainties.  
\begin{figure}[!h]
\includegraphics[width=\textwidth]{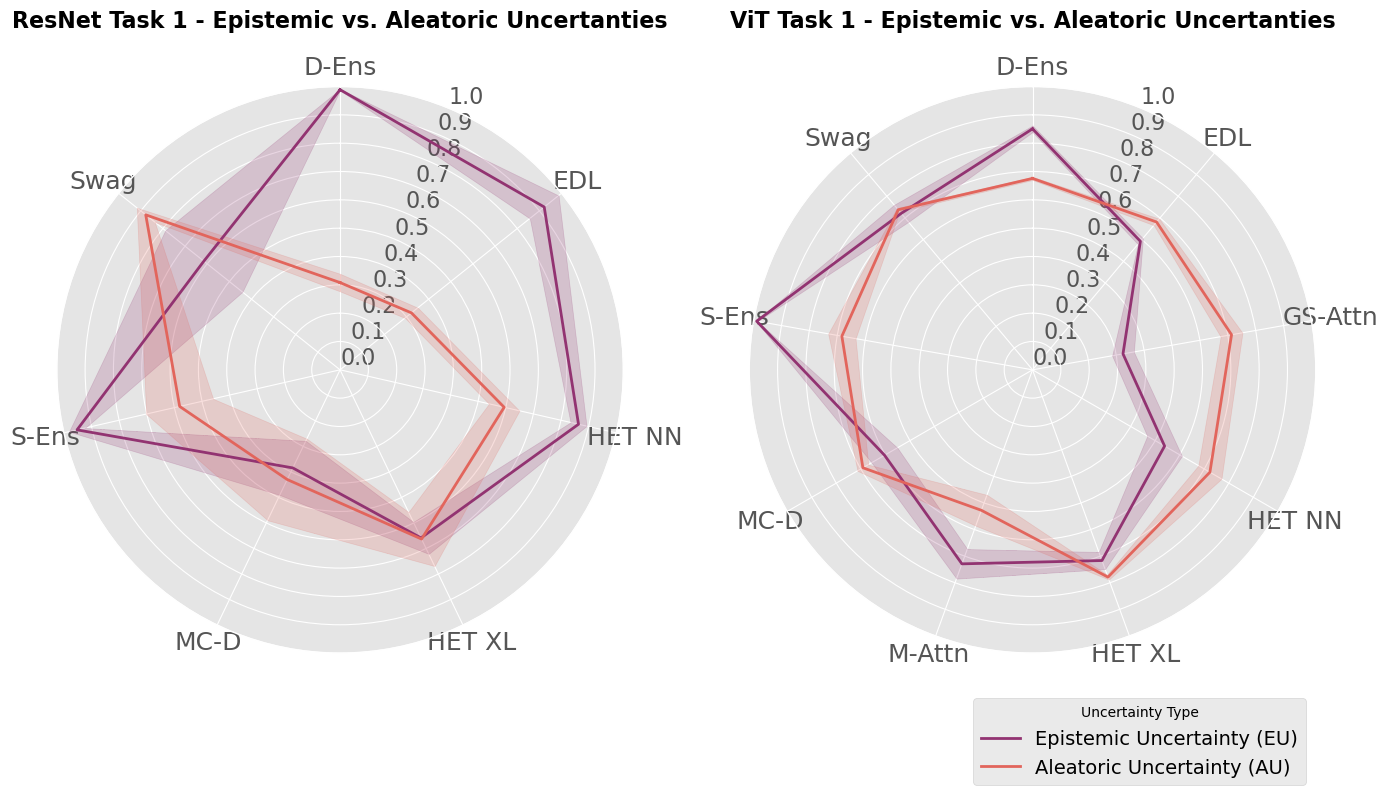}
\caption{Task 1 (OOD-Detection) AUROC scores for epistemic (purple) and aleatoric (red) uncertainty. Capability to disentangle is greater when epistemic uncertainty outperforms aleatoric uncertainty (i.e. purple line is further outwards than red line).} \label{fig_2:aleatoric_vs_epistemic}
\end{figure}
Given strong correlation between EU and AU scores —despite their intended non-correlation— we evaluate how well uncertainty estimates perform on their respective target tasks as final step of our analysis. We do this exemplary for Task 1 (OOD detection). Fig.\ref{fig_2:aleatoric_vs_epistemic} compares EU and AU estimators on Task 1. Only distributional methods are shown, as deterministic methods yield EU = 0 per \ref{sec:uc_decomposition}. Note that AUROC scores here do not necessarily align with AUROC scores of Task 1 in Fig.\ref{fig_3:method_results}, as it can also include AUROC derived from PU estimates, whereas in Fig.\ref{fig_2:aleatoric_vs_epistemic} we solely take EU and AU scores into account. Ideally, EU should significantly outperform AU in this task. However, many methods fail to disentangle uncertainty types, with AU estimates performing comparably or better. For ViT, only D-Ens, S-Ens, and M-Attn outperform aleatoric estimates. In ResNet, D-Ens, S-Ens, EDL, and Het-NN show significantly stronger epistemic performance. To synthesize our prior analyses, we aggregate metrics from Fig.\ref{fig_3:method_results} and Fig.\ref{fig_2:aleatoric_vs_epistemic} as well as rank correlations into an overview given in Fig.\ref{fig_4:overview}. Ensemble-based methods (D-Ens, S-Ens) consistently perform best across architectures, with HET-NN (ResNet) and M-Attn (ViT) coming closest in performance. Our results motivate several practical implications, discussed in the following section \ref{sec:conclusion}.

\begin{figure}[!ht]
\includegraphics[width=\textwidth]{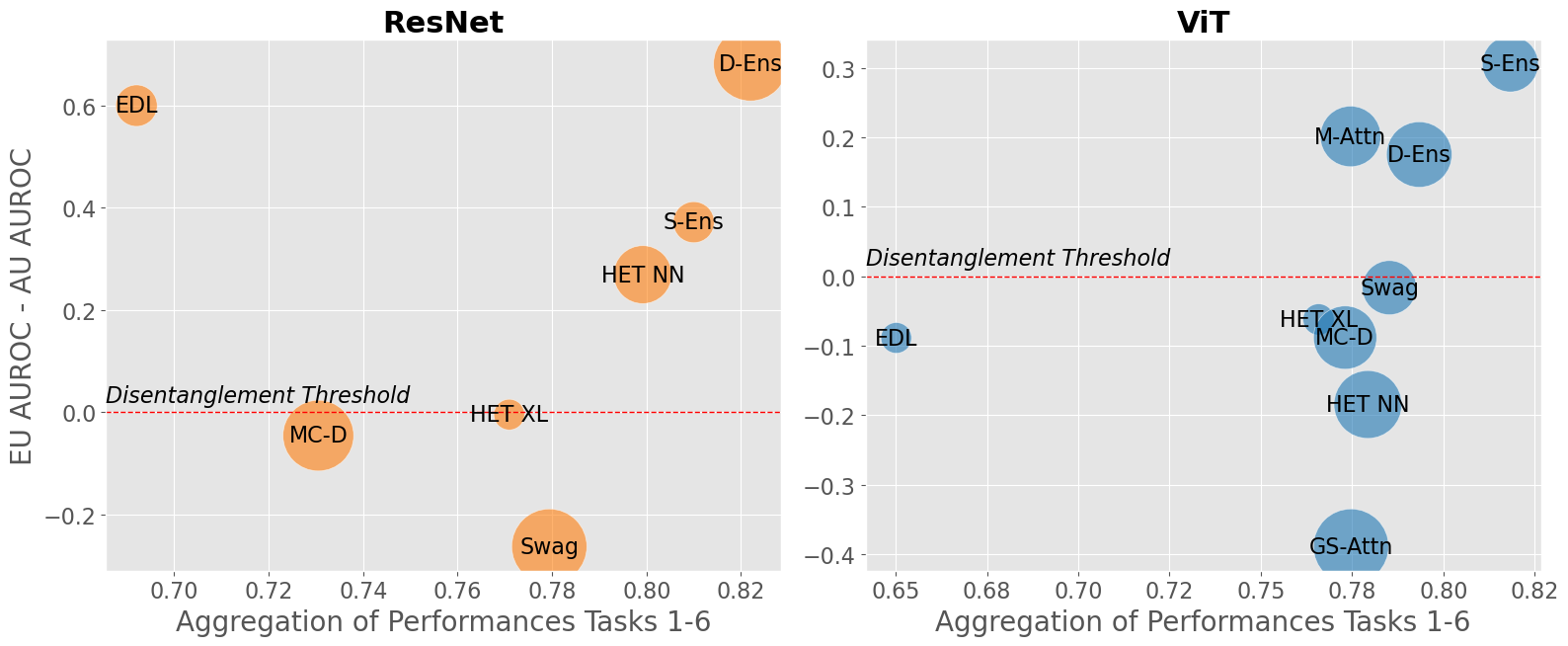}
\caption{Comparison of average performance across all tasks (x-axis; see Fig.~\ref{fig_3:method_results}) and disentanglement capability (y-axis), measured as $AUROC(\text{EU}) - AUROC(\text{AU})$ (see Fig.~\ref{fig_2:aleatoric_vs_epistemic}). Y-values above the disentanglement threshold indicate that EU and AU are well separated, with EU better aligned to the epistemic task. Values below the threshold suggest an inverse relationship, where AU unexpectedly performs better on the epistemic task. Bubble size reflects the inverse correlation between EU and AU scores; larger bubbles indicate weaker correlation. Methods in the upper-right region combine strong performance with well-separated uncertainty types.} \label{fig_4:overview}
\end{figure}

\section{Conclusion and Practical Implications}
\label{sec:conclusion}
We studied 13 UQ methods across two architectures—ResNet and ViT—and six tasks in a multilabel medical setting, establishing a comprehensive benchmark of UQ performance. Additionally, we assessed distributional methods' ability to decompose uncertainties using the information-theoretical approach. Based on our findings, we offer the following recommendations for clinical practitioners:

\subsubsection{Align models and methods with tasks.} We observed substantial variability in UQ method performance across tasks and architectures, highlighting the need to align the method with the specific UQ objective. For OOD detection, distributional (bayesian) methods clearly outperform deterministic ones and should be preferred. Notably, Swag with ResNet is highly effective for OOD, while performing poorly on aleatoric tasks—a potentially useful trait if understood. Conversely, for modeling aleatoric uncertainty, simple deterministic methods like LP and CP are surprisingly effective, performing nearly on par with ensembles.

\subsubsection{Architectures matter for calibration.} For well-calibrated predictions, practitioners should consider both the UQ method and the model architecture. ViTs generally outperform ResNets in calibration (MCE) across methods. SWAG remains a strong choice, while EDL should be used with caution, as it appears consistently ill-calibrated.

\subsubsection{Ensembles remain the most reliable.} Ensemble based UQ methods (D-Ens, S-Ens) consistently outperform other approaches across tasks while disentangling effectively in comparison and should therefore be preferred whenever possible.

\subsubsection{Understand Limits of IT-Based Disentanglement.} We showed that disentangling uncertainties via the information-theoretical approach can fall short of its theoretical expectations. If the IT framework holds, EU and AU should be uncorrelated and task-specific. However, in practice disentangled scores do not always align with their intended tasks—e.g., EU scores can counterintuitively perform worse on OOD detection than their AU and PU counterparts. We therefore recommend practitioners to apply the IT framework with caution and encourage  further research assessing limitations and practical reliability.

\subsubsection{Acknowledgments.}
This work was supported by the Senate of Berlin and the European Commision's Digital Europe Programme (DIGITAL) as grant TEF-Health (101100700).

This work was supported by the German Research Foundation (DFG) as research unit DeSBi [KI-FOR 5363] (459422098).

\subsubsection{Disclosure of Interests.}
The authors have no competing interests to declare that are relevant to the content of this article.

\newpage
\bibliographystyle{splncs04}
\bibliography{references}
\newpage
\section*{Appendix A}
\phantomsection
\label{sec:appendix_a}
\subsection*{Mutlilabel Adaptions}
\subsubsection{EDL}
Our multilabel adaption of EDL models uncertainty by treating predictions as samples from a Beta distribution \(\text{Beta}(\alpha_c^{(i)}, \beta_c^{(i)})\) instead of a Dirichlet Distribution, where $\alpha_c$ and $\beta_c$ represent the strength of evidence for a label. When the model is confident, these values are large, leading to low variance. Conversely, for uncertain predictions, $\alpha_c$ and $\beta_c$ remain small, increasing variance and uncertainty. The Kullback-Leibler(KL) divergence term regularizes this distribution, preventing overconfidence. This leads to the formulation of the EDL loss function:

\begin{align}
    L_{\text{Beta-EDL}} &= \frac{1}{n} \sum_{i=1}^{n} \sum_{c=1}^{C} 
    \left( \left( \frac{\alpha_c^{(i)}}{S^{(i)}} - y_c^{(i)} \right)^2 + \frac{\alpha_c^{(i)}}{S^{(i)}} \left(1 - \frac{\alpha_c^{(i)}}{S^{(i)}} \right) \frac{1}{S^{(i)} + 1} \right) \notag \\
    &\quad + \lambda_t D_{\text{KL}} \left( \text{Beta}(\alpha^{(i)}, \beta^{(i)}) \, || \, \text{Beta}(1,1) \right).
\end{align}

Here, \( y_c^{(i)} \) is the ground truth label for class \( c \) of sample \( i \), and \( \alpha_c^{(i)} \), \( \beta_c^{(i)} \) are the predicted Beta distribution parameters. The sum of evidence is \( S^{(i)} = \alpha^{(i)} + \beta^{(i)} \). The loss function comprises three components: the squared error between the Beta distribution mean and the target label, the variance of the Beta distribution, and a KL divergence regularization term. Uncertainty estimates for each label can be derived in closed form from  \(\alpha_c\) and \(\beta_c\). The total uncertainty PU is given by:

\begin{equation}    
\text{PU} = \psi(\alpha_c + \beta_c) - \frac{\alpha_c}{\alpha_c + \beta_c} \psi(\alpha_c) - \frac{\beta_c}{\alpha_c + \beta_c} \psi(\beta_c),
\end{equation}
where \(\psi(x)\) is the digamma function. Aleatoric uncertainty is calculated as:
\begin{equation}    
\text{AU} = -\frac{\alpha_c}{\alpha_c + \beta_c} \log\left(\frac{\alpha_c}{\alpha_c + \beta_c}\right) - \frac{\beta_c}{\alpha_c + \beta_c} \log\left(\frac{\beta_c}{\alpha_c + \beta_c}\right),
\end{equation}
epistemic uncertainty EU can be derived as $\text{EU} = \text{PU} - \text{AU}$. 

\subsubsection{Het Class NN}
In our multilabel adaptation of HetClassNN, the model predicts input-dependent heteroscedastic logit variances, modeling uncertainty for each label independently. Instead of softmax, we use sigmoid activation, and the loss is computed using binary cross-entropy (BCE) over Monte Carlo sampled logits:  

\begin{equation}
    \mathcal{L}_{\text{HetClassNN}} = \frac{1}{N} \sum_{i=1}^{N} \sum_{c=1}^{C} 
    \text{BCE} \left( y_c^{(i)}, \sigma(\mu_c^{(i)} + \sigma_c^{(i)} \epsilon) \right), 
    \quad \epsilon \sim \mathcal{N}(0,1),
\end{equation}  

where \( \mu_c^{(i)} \) and \( \sigma_c^{(i)} \) are the predicted mean and standard deviation for class \( c \), and \( \sigma(\cdot) \) is the sigmoid function.

\subsubsection{DDU}
In our multilabel adaptation of DDU, we derive uncertainty estimates post hoc by fitting a univariate Gaussian distribution to the features of each class independently. For each class \( c \), we begin by extracting the feature vectors of all samples labeled with that class. We then compute the class-specific mean \( \mu_c \) and variance \( \sigma_c^2 \) of these feature vectors to define the parameters of the Gaussian distribution. These parameters are calculated as follows:
\begin{equation}
\mu_c = \frac{1}{N_c} \sum_{i \in \text{class } c} x_i, \quad \sigma_c^2 = \frac{1}{N_c - 1} \sum_{i \in \text{class } c} (x_i - \mu_c)^2,
\end{equation}

where \( N_c \) is the number of samples belonging to class \( c \), and \( x_i \) is the feature vector of sample \( i \). To ensure the variance \( \sigma_c^2 \) remains positive, a small jitter \( \epsilon \) is added if necessary:

\begin{equation}
\sigma_c^2 \leftarrow \sigma_c^2 + \epsilon.
\end{equation}

Each class is thus modeled with a Gaussian distribution \( \mathcal{N}(\mu_c, \sigma_c^2) \), facilitating the estimation of prediction uncertainty by computing the negative log density during inference. 

\section*{Appendix B}
\phantomsection
\label{sec:appendix_b}
\subsection*{Rank Correlations of Epistemic and Aleatoric Uncertainties}
\begin{figure}[!h]
\includegraphics[width=\textwidth]{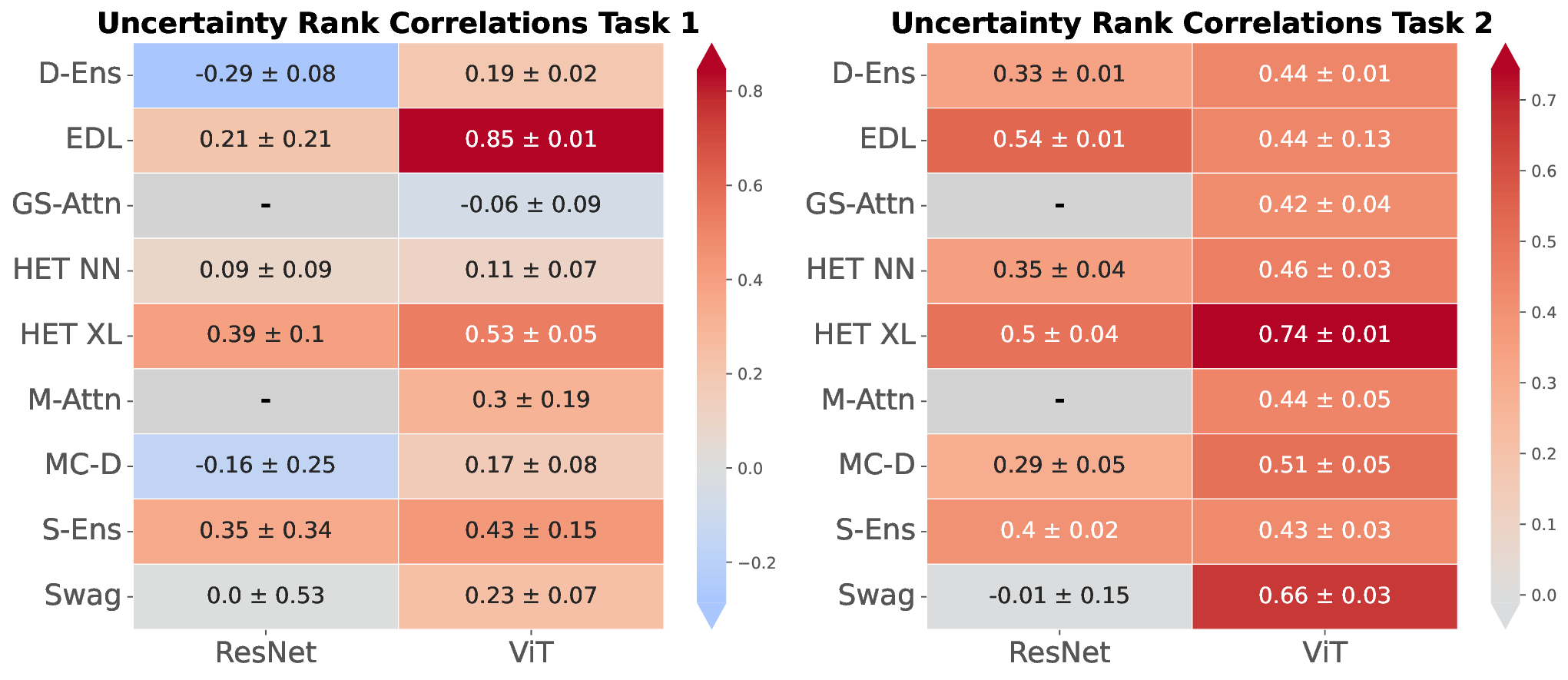}
\caption{Rank correlations between aleatoric and epistemic uncertainty scores for distributional methods on Tasks 1 and 2. For Task 1, ViT models exhibit high (> 0.5) (EDL, HET XL) and medium (0.1–0.5) correlations (D-Ens, M-Attn, MC-D, S-Ens, Swag, Het-NN), while GS-Attn shows slight negative correlation. ResNet models mostly show medium correlations (EDL, HET XL, S-Ens), while D-Ens and MC-D exhibit negative correlations. Swag varies widely across runs. In Task 2, ViT correlations remain above 0.4 and ResNet above 0.3, except for Swag ResNet, which shows no correlation. } \label{fig_1:correlations}
\end{figure}

\begin{comment}
\begin{figure}[!h]
\caption{Comparison of average performance across all tasks (x-axis; see Fig.~\ref{fig_3:method_results}) and disentanglement capability (y-axis), measured as $AUROC(\text{EU}) - AUROC(\text{AU})$ (see Fig.~\ref{fig_2:aleatoric_vs_epistemic}). Y-values above the disentanglement threshold indicate that EU and AU are well separated, with EU better aligned to the epistemic task. Values below the threshold suggest an inverse relationship, where AU unexpectedly performs better on the epistemic task. Bubble size reflects the inverse correlation between EU and AU scores (see Fig.~\ref{fig_1:correlations}); larger bubbles indicate weaker correlation. Methods in the upper-right region combine strong performance with well-separated uncertainty types.} \label{fig_4:overview}
\end{figure}
\end{comment}

\end{document}